\documentclass[10pt,twocolumn,letterpaper]{article}
\pdfoutput=1
\makeindex

\usepackage{cvpr}
\usepackage{times}
\usepackage{epsfig}
\usepackage{graphicx}
\usepackage{amsmath}
\usepackage{amssymb}
\usepackage{epstopdf}
\usepackage{dsfont}
\usepackage{color}
\usepackage[pagebackref=true,breaklinks=true,letterpaper=true,colorlinks,bookmarks=false]{hyperref}
\bgroup\catcode`\#=12\egroup  
\cvprfinalcopy % *** Uncomment this line for the final submission

\def\cvprPaperID{11} % *** Enter the wacv Paper ID here

\ifcvprfinal\fi
\begin{document}

%%%%%%%%% TITLE
\title{Diving deeper into mentee networks}

% Authors at the same institution
%\author{First Author \hspace{2cm} Second Author \\
% Authors at different institutions
\author{Ragav Venkatesan \\
Arizona State University\\
{\tt\small \href{mailto:ragav.venkatesan@asu.edu}{ragav.venkatesan@asu.edu}}
\and
Baoxin Li \\
Arizona State University\\
{\tt\small \href{mailto:baoxin.li@asu.edu}{baoxin.li@asu.edu}}\and
}

\maketitle
\ifcvprfinal\thispagestyle{empty}\fi

\begin{abstract}
Modern computer vision is all about the possession of powerful image representations. Deeper and deeper convolutional neural networks have been built using larger and larger datasets and are made publicly available. A large swath of computer vision scientists use these pre-trained networks with varying degrees of successes in various tasks. Even though there is tremendous success in copying these networks, the representational space is not learnt from the target dataset in a traditional manner. One of the reasons for opting to use a pre-trained network over a network learnt from scratch is that small datasets provide less supervision and require meticulous regularization, smaller and careful tweaking of learning rates to even achieve stable learning without weight explosion. It is often the case that large deep networks are not portable, which necessitates the ability to learn mid-sized networks from scratch.

In this article, we dive deeper into training these mid-sized networks on small datasets from scratch by drawing additional supervision from a large pre-trained network. Such learning also provides better generalization accuracies than networks trained with common regularization techniques such as $l_2$, $l_1$ and dropouts. We show that features learnt thus, are more general than those learnt independently. We studied various characteristics of such networks and found some interesting behaviors. 
\end{abstract}

\section{Introduction}
\label{sec:intro}
	%%%%%%%%%%%%%%%%%%%%%%%%%%%%%%%%%%%%%%%%%%%%%
	\begin{figure}[!t]
		\begin{center}
			\includegraphics[width=0.99\linewidth]{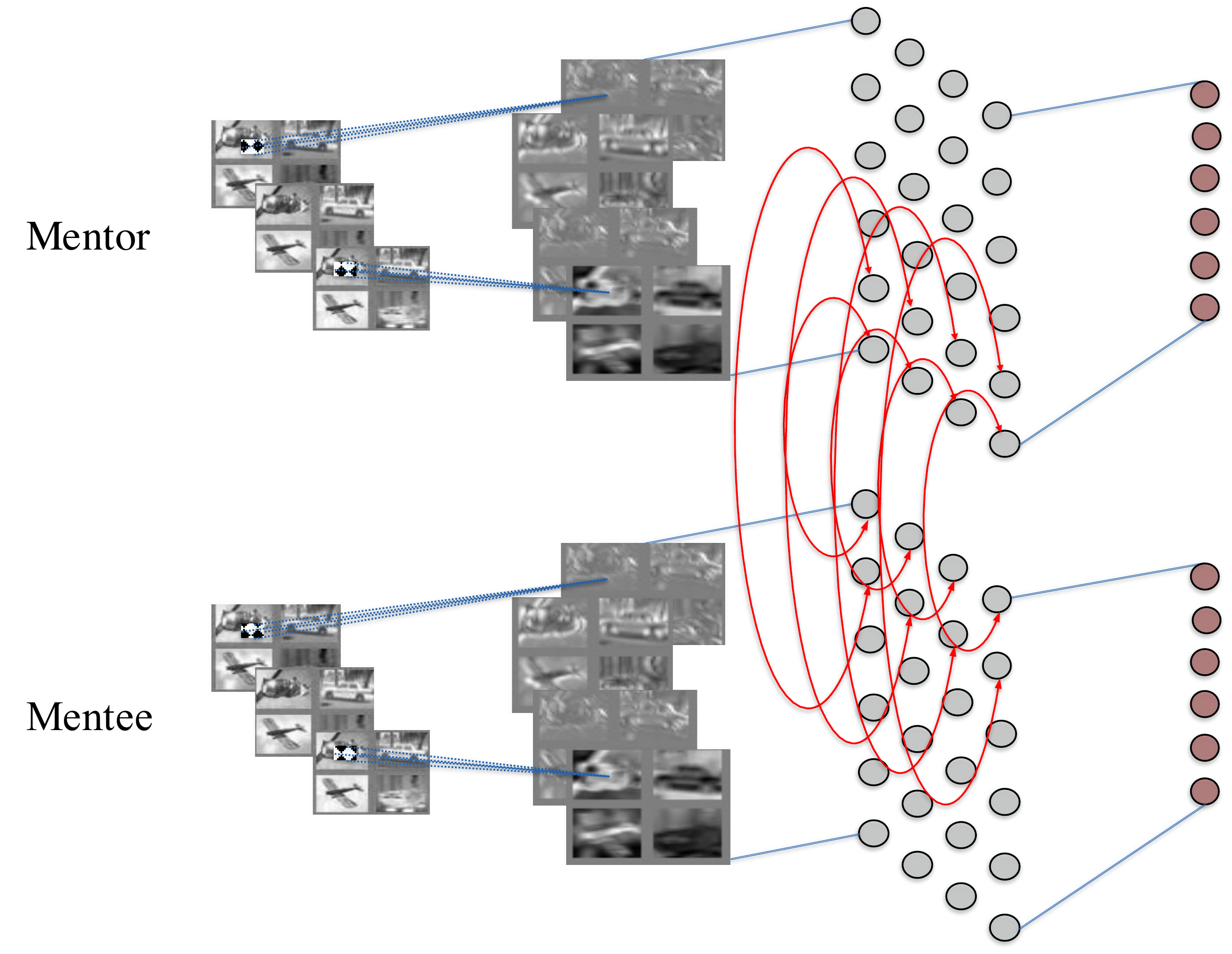}
		\end{center}
		\caption{Mentor mentoring mentee on the second hidden layer.}
		\label{fig:desc}
	\end{figure}
	%%%%%%%%%%%%%%%%%%%%%%%%%%%%%%%%%%%%%%%%%%%
With the proliferation of off-the-shelf, downloadable networks such as VGG-19, overfeat, R-CNN and several others in the caffe model zoo, it has become common practice in the computer vision community to simply fine-tune one of these networks for any task~\cite{SimonyanZ14a,jia2014caffe,girshick2014rich}. These networks are usually trained on a large dataset such as Imagenet and Pascal~\cite{ILSVRC15,everingham2010pascal}. The proponents of these networks argue that these networks have learnt image representations that are pertinent for most datasets that deal with natural images. Under the assumption that all these datasets are \emph{natural images} and are derived from a similar distribution this might as well be true. Even with such networks, features that are unique to each datasets do matter. While fine-tuning of an already trained network works to a certain extent, these features are not \emph{learnt} in a traditional manner on the target dataset but are simply copied. There is also no guarantee that these features are the best representations for the target dataset, although there is some validity in expecting that such a representation might work well, since after all it was learnt from a large enough dataset. 

Most computer vision scientists do not attempt to train a new architecture from scratch (random initializations). Training even a mid-sized deep network with a small dataset is a notoriously difficult task. Training a deep network, even those with mid-level depth require a lot of supervision in order to avoid weight explosion. On most imaging datasets, with image sizes being $224 X 224$,  the memory insufficiency of a typical GPU restricts the mini-batches to less than $100$. Using small mini-batches and small datasets lead to very noisy and untrustworthy gradients. This leads to weight explosions unless the learning rates are made sufficiently smaller. With smaller learning rates, learning is slowed down. With smaller mini-batches learning is unstable. One way to avoid such problems is by using regularization. By regularizing we can penalize the gradients for trying to make the weights go higher and higher. Batch Normalization is another technique that is quite commonly used to keep weight explosion under check~\cite{ioffe2015batch}. Even with these regularization techniques, the difficulty of training a deep network from scratch leads most computer vision scientists to use pre-trained networks.

There are several reasons why one might favour a smaller or a mid-sized network even though there might be a better solution available using these large pre-trained networks. Large pre-trained networks are computationally intensive and often have a depth in excess of $20$ layers. The computational requirement of these networks do not make them easily portable. Most of these networks require state-of-the-art GPUs to work even in simple feed forward modes. The impracticality of using pre-trained networks on smaller computational form factors necessitates the need to learn smaller network architectures. The quandary now is that smaller networks architectures cannot produce powerful enough representations.

Many methods have been recently proposed to draw additional supervision from large well-trained networks to regularize a new network while learning from scratch~\cite{wang2015recurrent,romero2014fitnets,balan2015bayesian,chan2015transferring}. All of these works were inspired from the Dark Knowledge (DK) approach~\cite{hinton2014dark}. All these techniques use at most one layer of supervision on top of the softmax supervision and try to use this technique to learn more deeper networks better. Figure~\ref{fig:desc} shows a conceptualization of this idea. 

In this paper, we try and make a shallower mentee network learn the same representation as a larger, well-trained mentor network at various depths of its network hierarchy. Mentorship happens by tagging on to the loss of the mentee network, a dissimilarity loss for each layer that we want mentored. To the best of our knowledge, there hasn't been any work that has regularized more than one layer this way. There also hasn't been any work that has trained a mid-sized network from a larger and deeper network from scratch. We study some idiosyncratic properties for some novel configurations of mentee networks. We argue that such mentoring avoids weight explosion. Even while using smaller mini-batches, mentee networks get ample supervision and are capable of stable learning even at high learning rates. We show that mentee networks produce a better generalization performance than an independently learnt baseline network. We also show that mentee networks are better transferable than the independently learnt baselines and are also a good initializer. We also show that mentee networks can learn good representations from very little data and sometimes even without supervision from a dataset in an unsupervised fashion.

The rest of the paper is organized as follows: section~\ref{sec:related} discusses related works, section~\ref{sec:proposed} details the mentored learning, section~\ref{sec:experiments} discusses designs for experiments, section~\ref{sec:results} produces results and section~\ref{sec:conclusions} provides concluding remarks.

\section{Related Works}
\label{sec:related}
Hinton et al., tried to make networks portable by learning the softmax outputs of a larger well-trained network along with the label costs~\cite{hinton2014dark}. This was previously explored using logits by Caruana et al.,~\cite{ba2014deep, buciluǎ2006model}. By directly learning the softmax layers, they were forcing the softmax layer of a smaller network to mimic the same mapping as that of a larger network onto the label space. In a way they tried to learn a better second and third guesses. They called this \emph{dark knowledge}, as the knowledge so learnt is only available to the larger network. By attempting to learn the softmax layer, they were able to transfer or \emph{distil} knowledge between the two networks. The drawback of this work is that it only works as long as the larger network is already well-trained and stable. They relied upon the network's predictive softmax layer being learnt perfectly on the target dataset and propagate that knowledge. This also assumes that there are relationships between classes to be exploited. While this may work in cases where this is true, such as in character recognition or in voice recognition, it doesn't work in most object detection datasets where the relationship between classes is not a given in terms of its appearance features\footnote{We tried this approach on Caltech101 and couldn't get reliable results.}. They also distil only the softmax labels and not the representational space itself. This also requires that the smaller network is capable of training in a stable manner. 

Dark knowledge is extended upon by several previous works~\cite{wang2015recurrent,romero2014fitnets,balan2015bayesian,chan2015transferring}. One extension of this work that we generalize in this article is using layer-wise knowledge transfer for one layer in the middle of the network. This was used to show that thinner and deeper network can be trained with better regularization~\cite{romero2014fitnets}. Another method uses a similar one-layer regularizer as knowledge transfer between a RNN and a CNN~\cite{chan2015transferring}. Mentored training has also been shown to be extremely useful when training LSTMs and RNNs with an independent mentor supervision~\cite{wang2015recurrent}.

All these methods discussed above are essentially the same technique as the dark-knowledge method extended beyond just the softmax layer. All of these methods have fixed one-layer regularizations and although trivial, we generalize this for many layers. Their mentee networks are typically much deeper and complex than their mentors and they use these as a means to build more complex models (albeit thinner as in the case of FitNets~\cite{romero2014fitnets}). There has been no study to the best of our knowledge that builds less complex (both thinner and shallower) models with the same capability as larger models. Also, neither has there been a study that studies various properties of these networks nor those that show the transferability and generality of these networks.

\section{Generalized mentored learning}
\label{sec:proposed}
		
Let us first generalize all of the methods that use this knowledge transfer as follows: Consider a large mentor network with $n$ layers $\mathcal{M}_n$. Suppose we represent the $k^{\text{th}}$ neuron activations of the $i^{\text{th}}$  layer in the network as $\mathcal{M}_n(i,k)$. Consider a smaller mentee network with $m$\footnote{Although we adhere strictly to $m<n$, without losing any generality, we could have any $m$ or $n$. In fact $m > n$ with only one probe would be the special case of FitNets~\cite{romero2014fitnets}.} layers $\mathcal{S}_m$. Suppose that $\mathcal{M}$ is already well-trained and stable on a general enough dataset $\mathcal{D}$. Now consider that we are using $\mathcal{S}$ to learn classification\footnote{Although we only consider the task of classification, the methods proposed are applicable to many forms of learning.} on a newer dataset $d_1$ which is less general and much smaller than $\mathcal{D}$ as determined a priori. Although this is not a constraint, having a smaller and less general dataset emphasizes the core need where such mentored learning is most useful. 

$\forall \ \ l\leq n$ and $ j \leq m$, we can define a probe as an error measure between $\mathcal{M}_n(l)$ and $\mathcal{S}_m(j)$. This error can be modelled as an RMSE error as follows,

\begin{equation}
\Psi(l,j) = 
\sqrt{\frac{1}{a}\sum_{i=0}^{a}(\mathcal{M}_n(l,i)-\mathcal{S}_n(j,i))^2},
\label{eqn:probe}
\end{equation}
where $a$ is the minimum number of neurons between $\mathcal{M}_n(l,.)$ and $\mathcal{S}(j,.)$. If the neurons were convolutional, we consider  element-wise errors between filters. By adding this cost to the label cost of the network $\mathcal{S}$  during back propagation, we learn not just a discriminative enough representation for the samples and labels of $d_1$, but also for layers $j$ in a pre-determined set of layers, a representation closer to the one produced by $\mathcal{M}$. Some implementations of such loses in literature tend to learn a regressor instead of simply adding the loss, but we concluded from our experiments that the computational requirements of such regressors do not justify their contributions. Adding a regressor would involve embedding the activations of the mentor and the mentee onto a common space and minimizing the distances between those embeddings. We quite simply circumvent that and consider the minimum number of matching neurons. This enables us to have a slimmer, fatter or same sized mentee. Suppose $d_1^{b}$ is the $b^{\text{th}}$ mini-batch of data from the dataset $d_1$ and suppose we have a pre-determined set of probes $B$, which is a set of tuples of layers from ($\mathcal{M}$, $\mathcal{S}$). The overall network cost is,
	
\begin{equation}
e = \alpha_t \mathcal{L}_s(d_1^{b}) + \beta_t \sum_{\forall (l,j) \in B} \Psi(l,j) + \gamma_t \Psi(n,m), 
\label{eqn:error}
\end{equation}	
where $\mathcal{L(.)}_s$ is the network loss of that mini-batch, $\alpha_t$ and $\beta_t$ weighs the two losses together to enable balanced training and $\gamma_t$ is the weight of the probe between the two (temperature) softmax layers. $\alpha_t = g_\alpha (t)$, $\beta_t = g_\beta(t)$ and $\gamma_t = g_\gamma(t)$ are annealing functions parametrized by the iteration $t$ under progress. Although most methods in the literature use constants for $\alpha_t, \beta_t$ and $\gamma_t$,  we found it preferable to retain $g_\alpha(t) = 1, \forall t$ throughout and anneal $\beta$ and $\gamma$ linearly. We discuss the value and the need for these parameters in detail further.

Since $\mathcal{M}$ is pre-trained and stable, the second and third terms of equation~\ref{eqn:error} are penalties for the activations of those layers in $\mathcal{S}$ not resembling the activations of the probed layer from $\mathcal{M}$ respectively. These losses as defined by equation~\ref{eqn:probe} are functions of the weights of those layers from $\mathcal{S}$ only. They restrict the weights within a proximity or region, that produces activations that are known for the mentor to be better activations. This restricting behaviour acts as a guided regularization process, allowing the weights to explore in a direction that the mentor thinks is a good direction, while still not letting the gradients to explode or vanish. 

For a particular weight $w \in \mathcal{S}$ at any layer, a typical update rule without the probe is,
\begin{equation}
w^{t+1} = w^{t} - \eta \frac{\partial}{\partial w} \mathcal{L}_s,
\label{eqn:norm_update}
\end{equation}
where $t$ is some iteration number, $\eta$ is the learning rate and assuming $\alpha_t = 1, \forall t$. The update rule with mentored probes is,
\begin{multline}
w^{t+1} = w^{t} - \eta \Big[ \alpha_t \frac{\partial}{\partial w} \mathcal{L}_s +  \\
 \beta_t \sum_{\forall (l,j) \in B} \frac{\partial}{\partial w}\Psi(l,j) + \gamma_t \frac{\partial}{\partial w} \Psi(n,m)  \Big].
\label{eqn:norm_update}
\end{multline}
The last two terms add a guided version of a \emph{noise} that decreases with each iteration. While at earlier stages of training, this allows the weights to explore the space, it also restricts the weights from exploding because the direction that the weights are allowed to explore is controlled by the mentor. The freedom to explore tightens up as the as learning proceeds, provided $g_\beta(t)$ is a monotonically annealing function with respect to $t$. Note that even though to calculate these error gradients we need one forward propagation through $\mathcal{M}$, we do not back propagate through $\mathcal{M}$. This is a penalty on the weights, even though we are using the activations to penalize the weights indirectly. Although mentee networks can be further regularized with $l_2$, $l_1$, dropouts and batch normalizations, it is recommended that the mentee networks imposes additional regularizations mirroring the mentor networks for better learning. 

\subsubsection*{Different configurations of mentee networks}
\label{sec:configs}
	
\begin{figure}[!t]
		\begin{center}
			\includegraphics[width=0.99\linewidth]{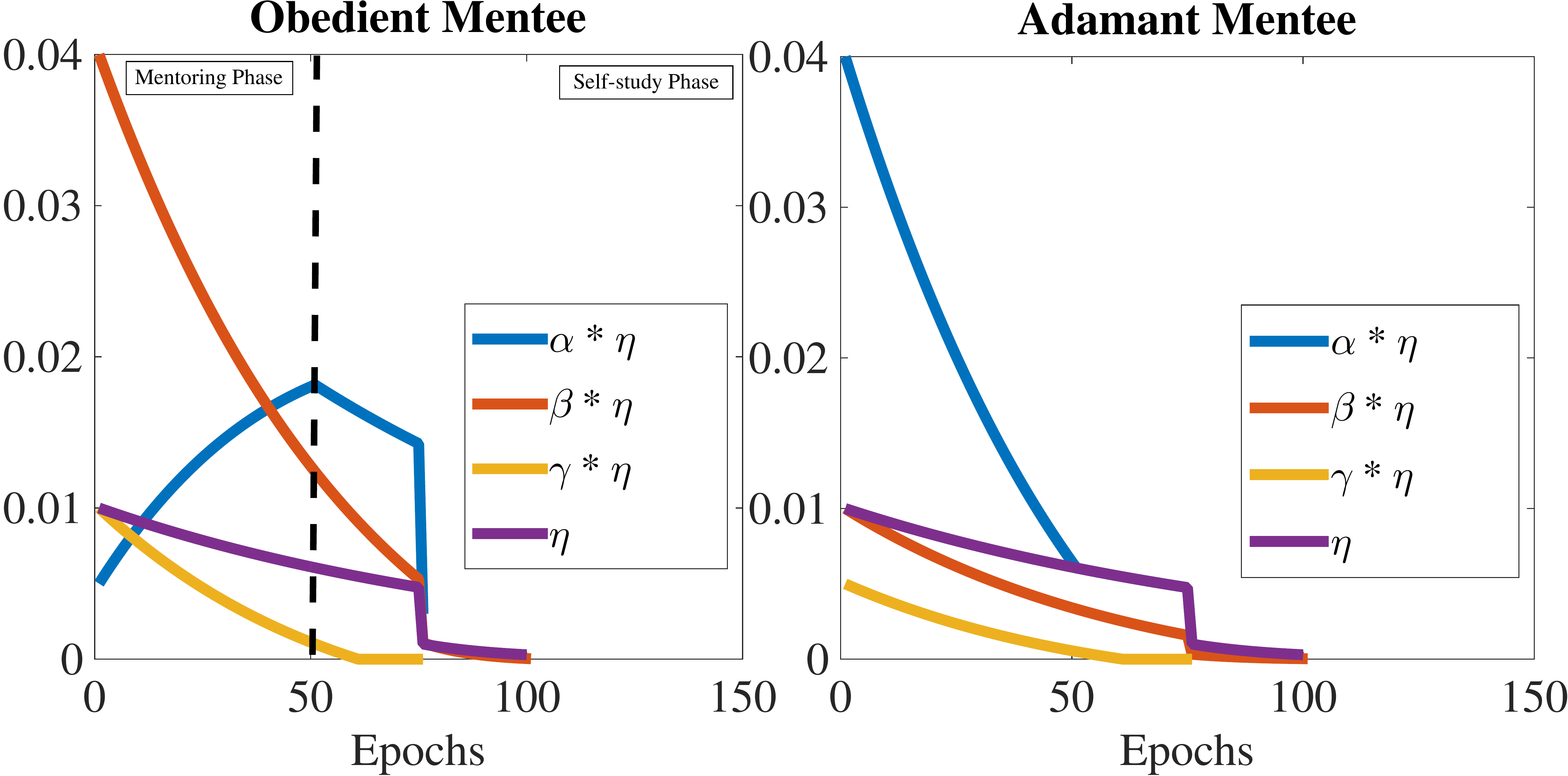}
		\end{center}
		\caption{Annealing $\alpha, \beta$ and $\eta$ while learning for an obedient and an adamant network.}
		\label{fig:students}
	\end{figure}

Different combinations of $\alpha$, $\beta$ and $\gamma$ produces different characteristics of mentee networks. Equation~\ref{eqn:norm_update} can be seen as learning with three different learning rates, $\alpha * \eta$, $\beta * \eta$ and $\gamma * \eta$. We can simulate using these three parameters, two idiosyncratic personalities of mentee networks: an obedient network and an adamant network. An obedient network is a network that focuses on learning the representation more than the label costs at the beginning stages and once a good representation is learnt, it focuses on learning the label space. It tends towards being over-regularized and its regularization relaxes with epochs. An adamant network is a network that focuses almost immediately on the labels as much as learning the representation, but its focus is positively towards learning the label only. The learning rates of these personalities are shown in figure~\ref{fig:students}.

	%%%%%%%%%%%%%%%%%%%%%%%%%%%%%%%%%%%%%%%%%%%%%
	%%%%%%%%%%%%%%%%%%%%%%%%%%%%%%%%%%%%%%%%%%%	
	
An independent network can be considered as a special case of the adamant network where probe weights are ignored ($\beta = 0$, $\gamma = 0,\ \forall t$). The other extreme case of an obedient network is perhaps a gullible network that learns just the embedding space of the mentor. Gullible networks are also a good way to initialize a network in an unsupervised mentoring fashion. Consider a dataset $d_2$, that does not have any labels. Neither the mentor nor the mentee could potentially learn any discriminative features. Using just the probes we could build an error function that could make the smaller mentee network still learn a good representation for the dataset. We use the information from the parent network to learn a good representation for $d_2$ by simply back propagating the second term of equation~\ref{eqn:error} alone. These gullible mentees come in really handy when the dataset has considerably less samples to be supervised with. Unsupervised mentoring is also an aggressive way to initialize a network and is often helpful in learning large networks in a stable manner with a stable initialization. 

Typically the deeper one goes, the more difficult it becomes to learn the activations and the costs saturate quickly. The softmax layer is the most difficult to learn. To our surprise we find that probe costs converge much sooner than the label costs, leading us to believe that the representations being mentored are indeed relevant as long as the datasets share common characteristics. There is a plethora of such configurations that could be tried and many unique characteristics discovered. In this article we limit ourselves to only those that enable us stability during learning and focus on those that help us with better generalizations.

For learning large networks we prefer the use of obedient networks as obedient networks are heavily regularized at the beginning leading to careful initialization and stabilization of the network before learning of labels takes over. We call the stabilization phase as the mentoring phase and the rest, self-study phase. During the mentoring phase learning is slow but steady. In most cases, $\alpha * \eta$ is an increasing function due to the aggressive climb of $\alpha$. The annealing of these rates for a typical obedient mentee and an adamant mentee are shown in figure~\ref{fig:students}. We also find that typically the later layers are more stubborn in being mentored than earlier layers. Although this is typically to be expected, more obedience may be enforced by choosing higher $\beta$ values for layers that are deeper in the network.

\section{Design of experiments}
\label{sec:experiments}	
						
We evaluate the effectiveness of mentorship through the following experiment designs:

\subsection{Effectiveness}

To demonstrate the effectiveness of learning, we first train a larger network on a dataset. Using this network as a mentor, we train the mentee network on the same dataset. Unlike those in literature, we choose mentee networks that are generally much smaller than the mentor. We show that this generalizes at least as well as an independent network of the exact same architecture regularized not by mentor, but by batch normalization, $l_2$ and $l_1$ norms and dropouts. Training mid-sized networks on small datasets are often difficult. To our best knowledge we have provided our best effort in meticulously learning all the networks. For learning an independent network often we spent additional effort in adjusting the learning rates at the opportune moments. We show that mentee networks outperforms the independent networks and even at the worst case performs as well as the independent networks.

\subsection{Generality of the learnt representations}
\label{sec:generality}
To demonstrate that the network learns a more general representation, we gather a pair of datasets of seemingly similar characteristics with one more general or larger than the other. We train the mentor with the more general dataset first and then fine tune it on the less general dataset. We then train both the independent and the mentee nets on the less general dataset and demonstrate again that at worst the mentee net performs the same as the independent net. 

We then proceed to fine tune the classifier layer of both the mentee net and the independent net using the more general dataset but since the other layers are not allowed to change, the mentee net does not have any additional supervision. This tests the quality of the features learnt by these networks on a more general and more difficult dataset. For the sake of our experiments we consider the pairing of (Cifar-10 - Cifar-100) and (Caltech-101 - Caltech-256)~\cite{krizhevsky2009learning,griffin2007caltech}. We assume that Cifar-100 is more general than Cifar-10 and Caltech-256 is more general than Caltech-101. 

Additionally, we conduct another experiment where we try to learn from a mentor network trained with the full MNIST dataset, a mentee network that only has supervision from a part of the dataset~\cite{lecun1998gradient}. The independent network also in this case, learns with the same redacted dataset. We redact the dataset by only having $p$ samples for each class in the dataset where $p \in \{ 500, 250, 100, 50 ,10, 1 \}$. $p = 1$ is essentially an ambitious goal of $1$-shot learning from scratch using a deep network. We also try this with a mentee network that is initialized by unsupervised mentoring from the same mentor network. We acknowledge that the comparison with unsupervised mentoring is unfair because the mentee net is initialized by the mentor with information filtered from data that is unavailable for the independent network. The latter results are to demonstrate that unsupervised mentoring could learn an effective feature space even without labels and with very less samples.

\subsection{Learning the VGG-19 representation}
	%%%%%%%%%%%%%%%%%%%%%%%%%%%%%%%%%%%%%%%%%%%%%
	\begin{figure}[!t]
		\begin{center}
			\includegraphics[width=0.99\linewidth]{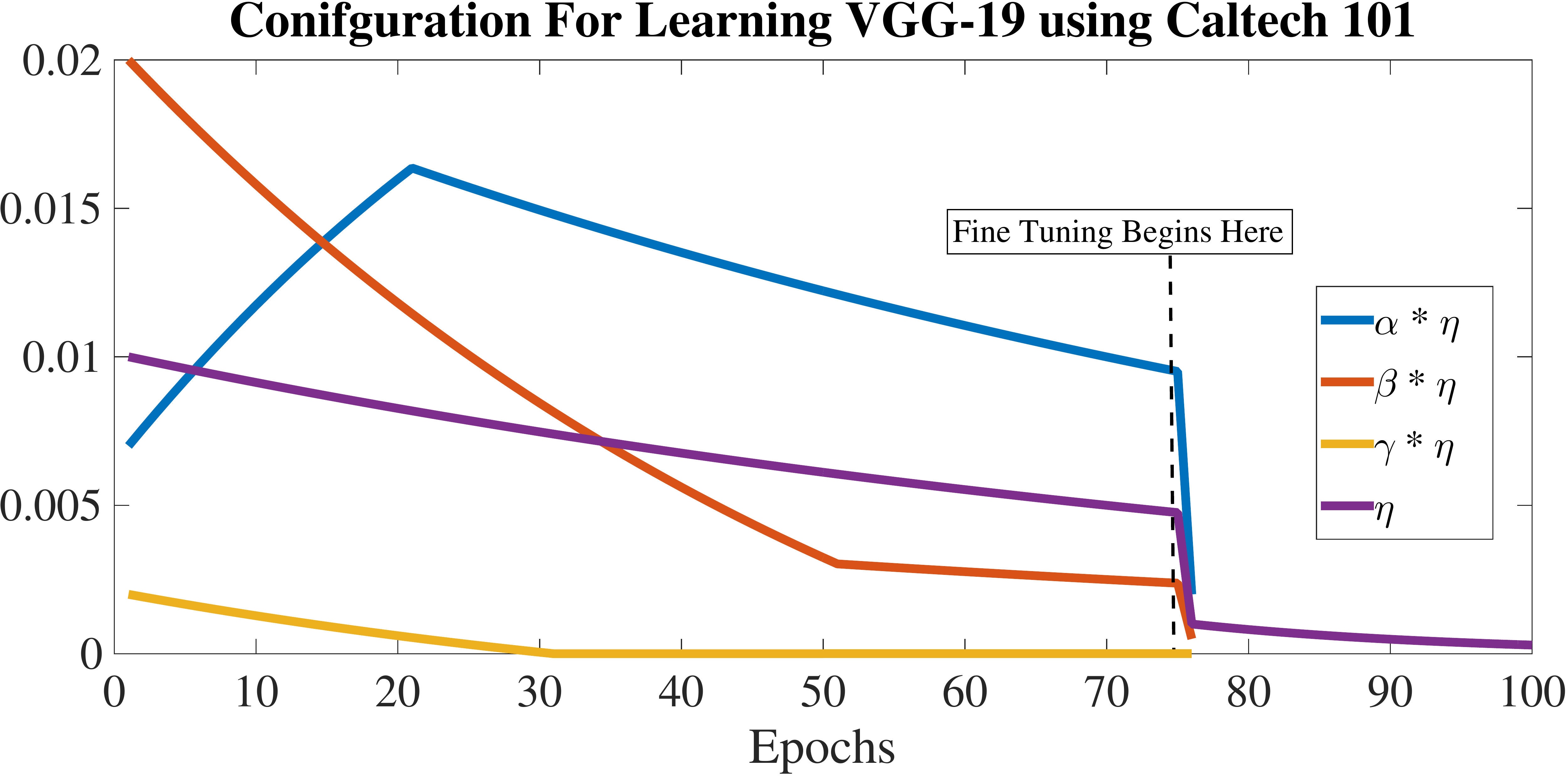}
		\end{center}
		\caption{Annealing $\alpha, \beta$ and $\eta$ while learning VGG-19 space for Caltech-101. We used an obedient network.}
		\label{fig:caltech101}
	\end{figure}
	%%%%%%%%%%%%%%%%%%%%%%%%%%%%%%%%%%%%%%%%%%%	
	
In particular, while learning classification on the Caltech101 dataset, we try to learn the same representation as the popularly used VGG-19 network at various levels of the network hierarchy~\cite{SimonyanZ14a}. VGG-19 network's $4096$ dimensional representation is one of the most coveted and iconic image features in computer vision at the time of the writing of this article. The VGG-19 network has $16$ convolutional layers and $2$ fully-connected layers the last of which produces the $4096$ dimensions of features upon which many other works have been built. 

We try to learn the same $4096$ dimensional representation of the VGG-19 network using ambitiously less number of  layers. For the (Caltech-101-Caltech256) dataset pairs in all our experiments, there is no explicit mentor network that we learnt. We simply set $g_\gamma(t) = 0, \forall t$ and learnt with probes without retraining the VGG-19 network. In a way we are attempting to learn VGG-19's view of the Caltech-101 dataset and are probing into the representational frame of the VGG-19 network. We used a relatively obedient student as shown in figure~\ref{fig:caltech101} for this case. 

\subsection{Implementation details}
The independent networks were all regularized with a $l_1$ and $l_2$ penalties with a weight of $1e^{-4}$, which seems to give the best results.  On all networks we also applied parametrized batch norm for both fully connected and convolutional layers and dropouts with rate of $p=0.5$ for the fully connected layers~\cite{ioffe2015batch,srivastava2014dropout}. We find that dropout and bath norm together help in avoiding over-fitting. All our activation functions were rectified linear units~\cite{nair2010rectified}. For learning the mentee network we start with learning rates as high as $0.5$, for the larger independent networks we are forced a learning rate of $0.001$, while for the smaller experiments we were able to go as high as $0.01$, since the batch sizes were larger. During training, if ever we ran into exploding gradients or NaNs, we reduce the learning rate by ten times, reset the parameters back to one epoch ago and continue training. We train until 75 epochs after which we reduce the learning rate by a hundred times and continue fine-tuning until early stopping. Unless early stopped, we train for $150$ epochs. All our initializations were from a $0$-mean Gaussian distribution, except the biases which were initialized with zeros. 

The experiment set-up was designed using \href{http://deeplearning.net/software/theano/}{Theano} v0.8 and the programs were written by ourselves\footnote{Code is available at our~\href{https://github.com/ragavvenkatesan/regularizer-network}{GitHub page}.}~\cite{Bastien-Theano-2012}. The experiments with MNIST datasets were conducted on a Nvidia GT 750M GPU, the others on an Nvidia Tesla K40c GPU, with \href{https://developer.nvidia.com/cudnn}{cuDNN} 3007 and \href{http://www.nvidia.com/object/cuda_home_new.html}{Nvidia CUDA} v7.5. The mini-batch sizes for all the MNIST and cifar experiments were 500 (unless forced by small dataset size in which case we performed batch descent instead of the usual stochastic descent). The mini-batch sizes for all Caltech experiments were 36, with images resized to $224X224$ so as to the fit the VGG-19 requirement. Apart from normalization and mean-subtraction, no other pre-processing were applied to any of the images. For the Caltech experiments we used Adagrad with Polyak's momentum~\cite{polyak1964some,green2013fast}. For the experiments that were smaller networks we used RMSprop with Nesterov's accelerated gradient~\cite{dauphin2015rmsprop,nesterov1983method}.

It is to be noted that we chose to use vanilla networks that are as simple as possible so as to enable us to compare against a baseline which is also vanilla. Since our aim is not to achieve state-of-the-art accuracies on any datasets, we didn't implement several techniques that are commonly applied to boost the network performances in modern day computer vision. The purpose of these experiments is to unequivocally demonstrate that among networks that learn from scratch, one that is mentored can perform better and learn more general features than one that is not.

\section{Results}
\label{sec:results}

	%%%%%%%%%%%%%%%%%%%%%%%%%%%%%%%%%%%%%%%%%%%%%
	\begin{figure*}[!t]
		\begin{center}
			\includegraphics[width=0.99\linewidth]{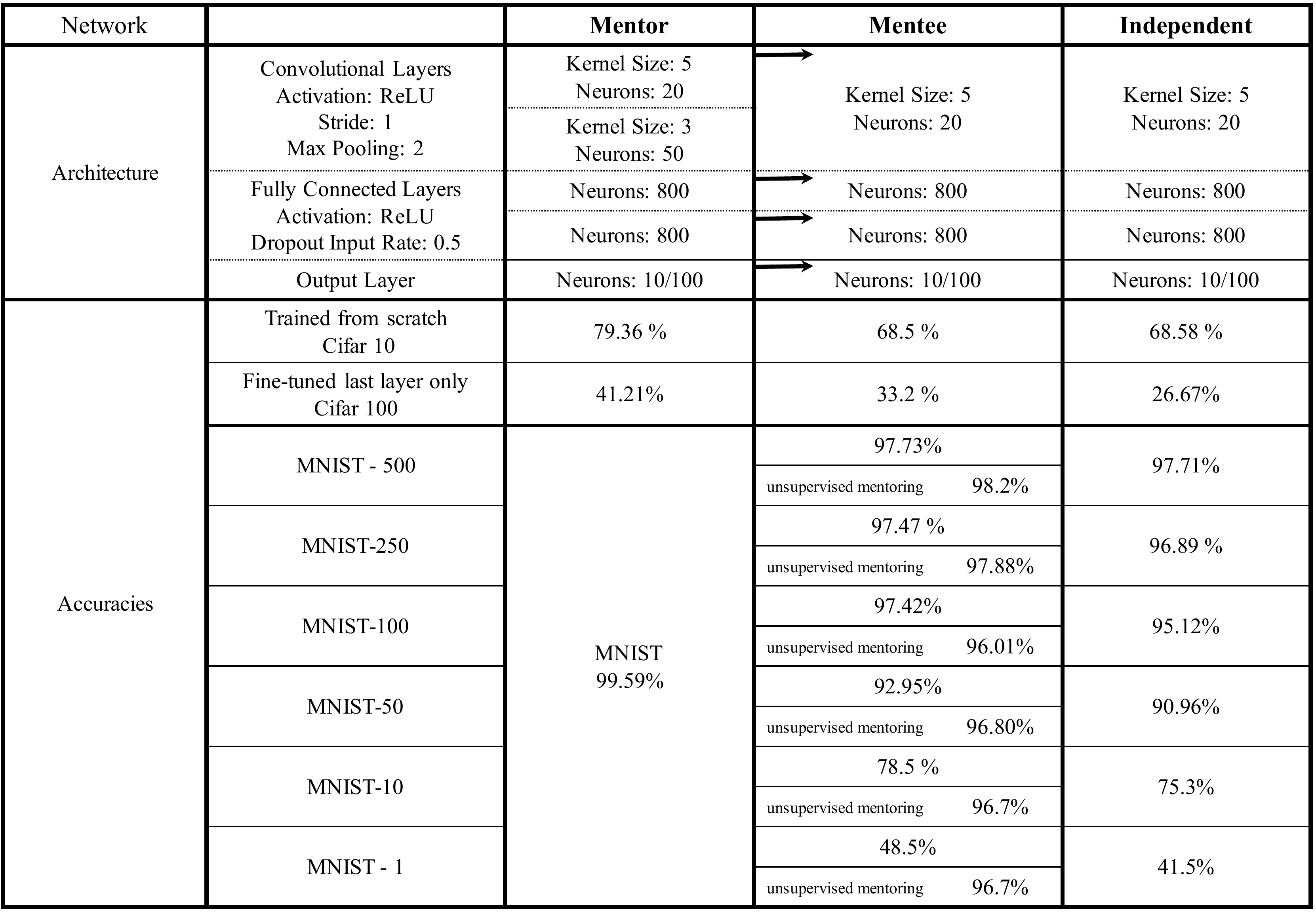}
		\end{center}
		\caption{Architecture and results for the experiments with CIFAR and MNIST datasets.}
		\label{fig:cifar-mnist-results}
	\end{figure*}
	%%%%%%%%%%%%%%%%%%%%%%%%%%%%%%%%%%%%%%%%%%
	
	%%%%%%%%%%%%%%%%%%%%%%%%%%%%%%%%%%%%%%%%%%%%%
	\begin{figure*}[!t]
		\begin{center}
			\includegraphics[width=0.99\linewidth]{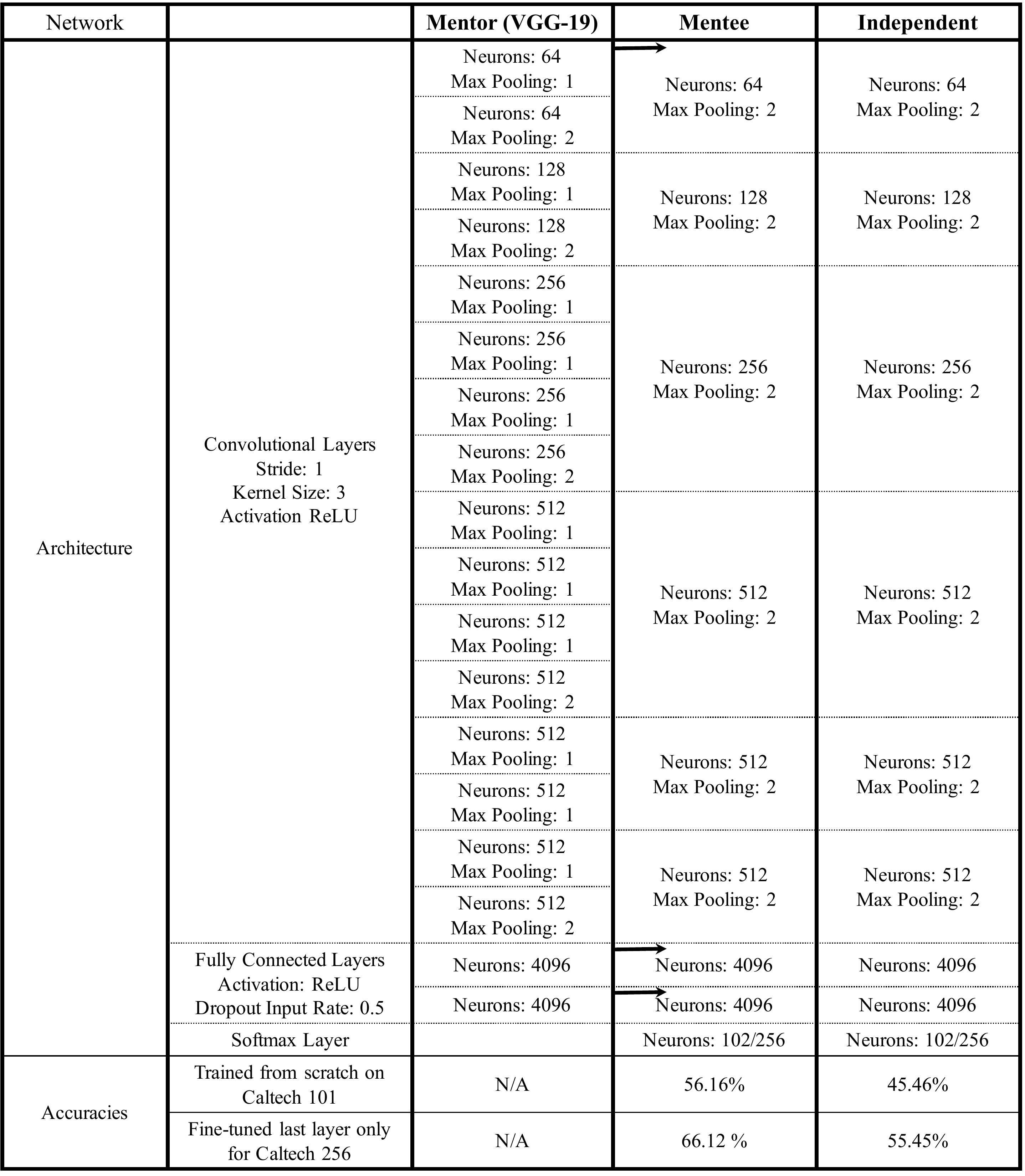}
		\end{center}
		\caption{Architecture and results for the experiments with Caltech datasets.}
		\label{fig:caltech-results}
	\end{figure*}
	%%%%%%%%%%%%%%%%%%%%%%%%%%%%%%%%%%%%%%%%%%
	
	%%%%%%%%%%%%%%%%%%%%%%%%%%%%%%%%%%%%%%%%%%%%%
	\begin{figure}[!t]
		\begin{center}
			\includegraphics[width=0.99\linewidth]{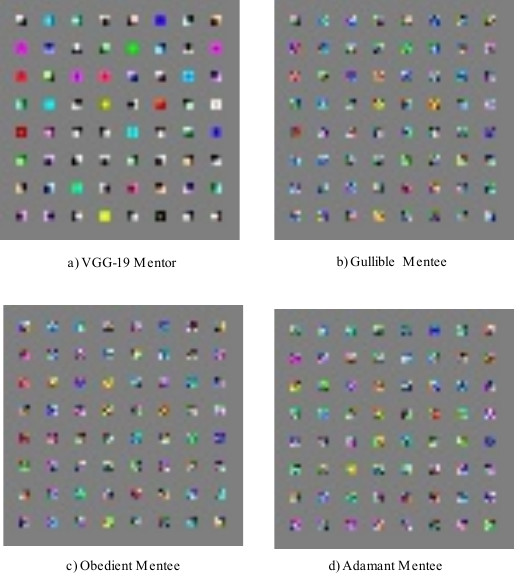}
		\end{center}
		\caption{VGG-19 first layer filters and filters probed using Caltech101 for a Gullible, Obedient and an Adamant mentee after only one epoch of training. We recommend viewing this image on a computer monitor.}
		\label{fig:filters}
	\end{figure}
	%%%%%%%%%%%%%%%%%%%%%%%%%%%%%%%%%%%%%%%%%%
		
The results are split across two tables based on the network architectures. The smaller experiments on a $5$ layer network are shown in figure~\ref{fig:cifar-mnist-results} and the larger ones in figure~\ref{fig:caltech-results}. The $\rightarrow$ symbol shows which layers are probed and from where.

In figure~\ref{fig:cifar-mnist-results}, the results clearly demonstrate the strong performance of the mentee networks over the independent networks. In the cifar experiments we under-weighted $\gamma$ purposely as we didn't want to propagate the $20\%$ of error from the mentor network on to the mentee network. The results on Cifar 10
from scratch seem to indicate that both networks have reached the best
possible performance for that architecture. We believe with the amount of supervision already provided from the 40,000 training images, mentoring is not
as effective. When there is already ample supervision, mentoring is ineffective,
or rather unwanted, albeit it doesn't hurt. While fine-tuning on cifar 100, we find that there are great
gains to be made.

We find a similar trend with the MNIST experiments also. The less data
there is, the higher the gain of the mentee networks. Note that even though mentee
networks are regularized, care was taken to ensure that they both go through the
exact same number of iterations at the exact same learning rate. We also found
that unsupervised mentoring always keeps the learning at a very high standard
although as was discussed in section~\ref{sec:generality} there was additional supervision on the
entire dataset from the unsupervised mentoring, which is unfair.

In the experiments with the Caltech101 datasets, we find that the mentee
networks perform better than the vanilla network. The
mentee network was also able to perform significantly better than the independent
network when only the classifier/mlp sections were allowed to learn the Caltech256 dataset with representation learnt from Caltech101. This proves the
generality of the feature space learnt. With an even obedient student, we were
able to learn the feature space of the VGG-19 network to a remarkable degree.
While with the first convolutional layer we were able to learn to a minimum
rmse or $0.0023$ from $6.54$ at random. With the last two layers we were able to
learn upto a rmse of $2.04$ from $12.76$ at random.

Figure~\ref{fig:filters} shows the filters learnt after one epoch for a gullible network, an  obedient network and an adamant network. All these networks were initialized with same random values at their inception. We can easily notice that the gullible network already sway towards the VGG-19 filters. In obedient mentee, we notice that most corner detector features are already swaying towards the mentee network but more complex features are not swaying as much as the gullible network. To our surprise we notice that even in an adamant network corner detectors are swaying towards VGG-19. This shows that even with low weights, the first layer features are learning the VGG-19's representation. It is to be noted that we are not learning the weights directly, but are learning the activations produced by the VGG-19 network for the Caltech101 dataset that leads us to learn the same filters as the VGG-19. This implies that corner features are more general among the Imagenet dataset, which VGG-19 was trained on, and the Caltech101 dataset, which explains why they are learnt earlier than others.

\section{Conclusions}
\label{sec:conclusions}

While the use of large pre-trained networks will continue to remain popular,
because of the ease in just copying a network and fine-tuning the last layers,
we believe that there is still a need for learning small and mid-sized networks
from scratch. We also recognize the difficulty involved in reliably training deep
networks with very few data samples. One way to meet the best of
both worlds is by using a mentored learning approach. In our study, we find that a shallower 
mentee network was able to learn a new representation from scratch while being 
regularized by the mentor network's activations for the same input samples.
We found that such mentoring provided much stabler training even at higher
learning rates. We noted some special cases of these networks and recognize some idiosyncratic
personalities. We extended one of these to be able to perform as an unsupervised 
initialization technique. We showed through compelling experiments, the strong
performance and generality of mentor networks.

{\small	
\bibliographystyle{ieee}
\bibliography{arxiv}
}

\end{document}